\documentclass{article}
\usepackage{array}

\usepackage[preprint]{neurips_2026}
\usepackage{graphicx}
\usepackage{amsmath}
\usepackage{multirow}
\usepackage{float}
\usepackage{cuted}
\usepackage{capt-of}
\usepackage{booktabs}
\usepackage{colortbl}

\usepackage[utf8]{inputenc} %
\usepackage[T1]{fontenc}    %
\usepackage{hyperref}       %
\usepackage{url}            %
\usepackage{booktabs}       %
\usepackage{amsfonts}       %
\usepackage{nicefrac}       %
\usepackage{microtype}      %
\usepackage{xcolor}         %
\newcolumntype{C}[1]{>{\centering\arraybackslash}p{#1}}

\title{\textsc{MOF-Verify}: A Failure-Aware Agentic Harness for MOF Hypothesis Verification}

\author{%
  \mdseries
  \textbf{Donghyun Lee}$^{1}$\thanks{These authors contributed equally.}\;
  \textbf{Taehoon Lee}$^{2}$\footnotemark[1]\;
  \textbf{GeonHee Ahn}$^{1}$\footnotemark[1]\;
  \textbf{Jieun Kim}$^{2}$\footnotemark[1] \\
  Jihyun Park$^{1}$\;
  Suyeon Cho$^{1}$\;
  Yoona Kim$^{1}$\;
  Chaerim Shin$^{1}$ \\
  Hoi Ri Moon$^{1}$\thanks{Corresponding authors. E-mail:
  \texttt{\{hoirimoon,jgna,sookim\}@ewha.ac.kr};
  \texttt{\{sh,jihwan\}@lymeric.ai}.}\;
  Jonggeol Na$^{1}$\footnotemark[2]\;
  Sukho Hong$^{2}$\footnotemark[2]\;
  Jihwan OH$^{2}$\footnotemark[2]\;
  Soo Kyung Kim$^{1}$\footnotemark[2] \\
  $^{1}$Institute for Multiscale Matter and Systems (IMMS), \\
  Ewha Womans University, Seoul, Republic of Korea \\
  $^{2}$Lymeric, Seoul, Republic of Korea
}

\begin{document}

\maketitle

\begin{abstract}
Large language models are increasingly used as reasoning components in AI-driven materials Co-Scientists, yet the reliability of the resulting verification pipeline remains unclear. Metal--organic frameworks (MOFs) provide a particularly challenging setting because structures may appear under different identifiers, synthesis outcomes depend strongly on experimental conditions, evidence is distributed across heterogeneous sources, and some hypotheses require computation rather than literature alone. We introduce a diagnostic benchmark with four task families covering structural grounding, synthesis-condition verification, evidence-sufficiency verification, and MLIP-based computational verification. T-MOF-1--3 are evaluated under closed-book, retrieval-enabled, and oracle-evidence settings to localize failures in knowledge access, evidence acquisition, and reasoning, while T-MOF-4 separately evaluates computational verification. Guided by these diagnosed failure modes, we develop \textsc{MOF-Verify}, a failure-aware agentic harness that targets structural, literature, evidence-sufficiency, and computational bottlenecks before producing a final verdict. Across multiple backbone LLMs, \textsc{MOF-Verify} improves the reported average over direct inference and retrieval-based baselines on this benchmark, with gains concentrated in structural grounding and evidence sufficiency. Benchmark records and release-status documentation are available at \url{https://github.com/IMMS-Ewha/MOF-Verify-Benchmark}.
\end{abstract}

\section{Introduction}
\label{sec:introduction}

Metal--organic frameworks (MOFs) are a highly tunable class of materials with applications in gas separation, carbon capture, catalysis, and energy storage~\citep{furukawa2013chemistry,li2012mofseparations,sumida2012co2capture}. 
Their large design space, created by combining metal nodes with organic linkers, enables fine control over structure and function but also makes discovery difficult to navigate manually~\citep{furukawa2013chemistry,zhou2012introduction}. 
Relevant knowledge is distributed across a rapidly growing literature, structural databases, and computational studies, making MOFs a natural domain for AI-assisted discovery~\citep{wilmer2012largescale,chung2014core,batra2021emerging}. 
Recent LLM-based and agentic systems further expand this possibility by connecting scientific literature with external databases and computational tools~\citep{bran2024chemcrow,boiko2023autonomous}. 
Accordingly, LLMs and agents have been explored for MOF information extraction, synthesis planning, screening, simulation, and automated discovery~\citep{zheng2023chatgpt,kang2025harnessing,pruyn2025mofchemunity,lee2026simmof,ahn2026mind}.

However, reliable MOF verification requires substantially more than generating a plausible answer from an LLM. 
The same or related framework may appear under different names or structural identifiers~\citep{pruyn2025mofchemunity}; synthesis and measured properties can depend strongly on experimental conditions~\citep{agrawal2020repeat,sholl2026replicate}; and evidence from different studies may not be directly comparable because of differences in structure, synthesis protocol, or computational setup. 
Moreover, some hypotheses cannot be settled from literature evidence alone and require a scientific calculation before a verdict can be made. 
Reliable MOF hypothesis verification therefore involves a chain of judgments spanning material grounding, experimental-context recovery, evidence acquisition, recognition of insufficient evidence, and, when necessary, computational verification.

This broader pipeline makes it important to identify \emph{where} and \emph{why} a MOF Co-Scientist fails. 
An incorrect verdict may originate from retrieval of the wrong structural evidence, recovery of incorrect synthesis information, failure to recognize that the available evidence is insufficient, or failure of a computational tool to reliably resolve the requested quantity. 
These failures require different interventions but are easily conflated by end-to-end accuracy alone. 
We therefore introduce a diagnostic benchmark that exposes failures at distinct stages of the MOF verification pipeline, and use the resulting diagnosis to construct \textsc{MOF-Verify}, a verification harness that explicitly addresses the identified bottlenecks before producing a final verdict.

In this work, we study two research questions.

\textbf{RQ1: Where and why does a MOF Co-Scientist fail in hypothesis verification?}
To systematically analyze verification failures, we construct a diagnostic benchmark with four task families: \textbf{T-MOF-1} tests structural retrieval and grounding; \textbf{T-MOF-2} tests synthesis-information retrieval and verification; \textbf{T-MOF-3} tests evidence-sufficiency verification; and \textbf{T-MOF-4} tests MLIP-based computational verification. 
T-MOF-1--3 are evaluated under three controlled evidence-access settings---\textbf{closed-book}, using only the LLM's parametric knowledge; \textbf{retrieval-enabled}, with external evidence retrieved through RAG; and \textbf{oracle-evidence}, where the evidence required for verification is directly provided. 
Across these settings, the system predicts \textit{Yes}, \textit{No}, or \textit{Uncertain}. 
The three task families and three evidence-access settings form a $3\times3$ diagnostic matrix that allows us to localize both \emph{where} a verification failure occurs and \emph{why} it emerges as evidence access changes. 
T-MOF-4 separately compares Closed-book with computational verification: the Co-Scientist executes an MLIP-based calculation~\citep{jacobs2025practical}, yielding \textit{Yes} when a valid result is obtained, \textit{Re-run} when the calculation does not converge, and \textit{Human-review} when the requested quantity falls outside what the calculator can determine.

\textbf{RQ2: Can a failure-aware agentic harness mitigate the identified MOF verification bottlenecks?}
Using the diagnostic results, we build \textsc{MOF-Verify} around a frozen base LLM. \textbf{Identity Resolution (IR)} resolves MOF names and aliases to a canonical material identity; \textbf{Structural Evidence (SE)} retrieves and validates the corresponding trusted deposited structure; \textbf{Literature Evidence (LE)} checks synthesis conditions against the paper's main text and supporting information; \textbf{Evidence Sufficiency (ES)} assesses whether the evidence supports a conclusion despite conflicting findings or missing conditions; and \textbf{Computational Evidence (CE)} executes requested calculations and checks their convergence and scope. \textbf{Verdict Routing (VR)} integrates the task-relevant outputs to assign the final verdict or abstain. LE and ES use the LLM, while IR, SE, CE, and VR are rule-based. Section~4 describes the task-specific decision paths. Evaluation on the same benchmark measures the combined system, not isolated module effects.

Our key contributions are threefold:
\begin{itemize}
    \item We introduce an expert-informed diagnostic benchmark with four task families covering structural retrieval, synthesis-information retrieval, evidence-sufficiency verification, and MLIP-based computational verification.
    \item We propose a controlled $3\times3$ evaluation design for T-MOF-1--3, together with a separate computational protocol for T-MOF-4, to localize \emph{where} and \emph{why} a MOF Co-Scientist fails.
    \item Guided by these diagnosed failure modes, we develop \textbf{\textsc{MOF-Verify}}, a failure-aware agentic harness comprising \textbf{Identity Resolution }, \textbf{Structural Evidence }, \textbf{Literature Evidence}, \textbf{Evidence Sufficiency}, and \textbf{Computational Evidence}, with a shared \textbf{Verdict Routing} module that integrates their outputs and determines the final verdict.
\end{itemize}

\section{Related Work}
\label{sec:related_work}

LLMs and agentic systems are increasingly used to automate scientific workflows in materials science. General-purpose agents such as ChemCrow~\citep{bran2024chemcrow} and Coscientist~\citep{boiko2023autonomous} augment LLMs with external tools for planning and executing chemistry tasks, while Google DeepMind's Co-Scientist~\citep{gottweis2026coscientist} extends this paradigm to multi-agent hypothesis generation with experimental validation. Closed-loop systems such as A-Lab~\citep{szymanski2023alab} similarly combine computational screening with autonomous synthesis. For MOFs, prior work has used LLMs to extract synthesis conditions and properties from the literature~\citep{zheng2023chatgpt,kang2025harnessing,pruyn2025mofchemunity}, automate simulation workflows~\citep{lee2026simmof}, and couple hypothesis generation with computational validation using MLIPs~\citep{ahn2026mind}. Together with large-scale screening~\citep{wilmer2012largescale} and structural databases~\citep{chung2014core}, these efforts demonstrate the potential of LLM-based Co-Scientists to connect literature, structured materials data, and computational tools throughout the MOF discovery process.

Alongside these systems, scientific benchmarks have been developed to evaluate LLM knowledge and reasoning. GPQA~\citep{rein2023gpqa} and MaScQA~\citep{zaki2023mascqa} assess scientific reasoning across physics, chemistry, and materials science, while MOF-specific benchmarks evaluate knowledge, literature extraction, database querying, property prediction, experiment design, and tool-augmented generation~\citep{bai2024evaluation,an2023kgqa4mat,kang2024chatmof}. These benchmarks establish important capabilities, but they largely evaluate final task performance rather than tracing an incorrect scientific verdict to a specific stage of a Co-Scientist pipeline.

This distinction becomes critical when verification depends on multiple components. Retrieval-augmented generation~\citep{lewis2020rag} may fail because the relevant evidence is not retrieved or because retrieved evidence is incorrectly interpreted~\citep{jin2025disentangling}; in MOF workflows, additional failures can arise from ambiguous material identity, condition-dependent evidence, invalid cross-study comparison, or an external computational tool that cannot reliably settle the hypothesis. Existing MOF-oriented systems and benchmarks~\citep{bai2024evaluation,an2023kgqa4mat,kang2024chatmof,zheng2023chatgpt,kang2025harnessing,lee2026simmof,ahn2026mind} do not systematically localize these failures across the full verification pipeline. Our work addresses this gap with a controlled diagnostic benchmark that identifies \emph{where} and \emph{why} a MOF Co-Scientist fails, and \textsc{MOF-Verify}, a failure-aware agentic harness designed to directly compensate for the resulting bottlenecks.

\section{Diagnostic Benchmark}

\subsection{Benchmark Design for MOF Hypothesis Verification}

MOF hypothesis verification can fail at several distinct stages: a Co-Scientist may (1) retrieve the wrong structural evidence, (2) recover incorrect synthesis information, (3) overlook that the available evidence is insufficient to support a unique conclusion, or (4) fail to obtain the computational evidence required to resolve the claim. We therefore construct four corresponding task families, with \textbf{T-MOF-1--4} consisting of hypothesis--answer pairs designed to isolate failure modes (1)--(4), respectively. \textbf{T-MOF-1--3} evaluate evidence-based verification under three controlled evidence-access settings---\textbf{closed-book}, \textbf{retrieval-enabled}, and \textbf{oracle-evidence}---whereas \textbf{T-MOF-4} separately evaluates whether the Co-Scientist can obtain the computational evidence required to resolve a hypothesis through an MLIP-based tool-execution protocol. The composition of T-MOF-1--4, including the number of items, controls, abstention cases, and evidence anchors, is summarized in Table~\ref{tab:dataset}.

\begin{table}[t]
  \centering
  \small
\caption{Composition of the T-MOF-1--4 diagnostic benchmark. We report the number of items, adversarial control cases, abstention-labeled examples, and evidence anchors for each task family. }
  \label{tab:dataset}
  \begin{tabular}{lrrrl}
    \toprule
    Task & Items & Controls & Abstention gold & Evidence anchor \\
    \midrule
    T-MoF-1 & 276 & 130 (47\%) & 6  & CSD refcode \\
    T-MoF-2 & 869 & 425 (49\%) & 53 & DOI + verbatim span \\
    T-MoF-3 & 117 & 74 (63\%)  & 18 & DOI + decisive condition \\
    T-MoF-4 & 8   & 0 (0\%)    & 5  & MLIP calc.\ spec.\ hash \\
    \midrule
    Total   & 1{,}270 & 629 (50\%) & 82 & \\
    \bottomrule
  \end{tabular}
\end{table}

\textbf{T-MOF-1: Structural Identity.}
T-MOF-1 evaluates whether the system can retrieve and verify the correct crystal structure for a material described by a name, alias, or formula in the literature. Items are constructed from curated identity records linked to CSD refcodes, such that answering correctly requires grounding the material to the appropriate deposited structure. We additionally include deliberately confusing controls to prevent the task from becoming a simple lookup exercise. In particular, decoys use real and plausible CSD refcodes that actually belong to a different framework. The system must therefore verify that the retrieved structural record truly corresponds to the intended MOF rather than merely recognize a familiar material name. T-MOF-1 consequently measures the reliability of CSD-based structural retrieval and grounding.

\textbf{T-MOF-2: Synthesis-condition Verification.}
T-MOF-2 evaluates whether the system can recover the correct synthesis conditions reported for a MOF in the primary literature. Each item specifies a synthesis route that may include temperature, time, solvent, modulator, metal source, and linker source, and the system must determine whether the cited paper actually reports that route. Gold evidence is taken verbatim from the primary literature and paired with the corresponding DOI. Negative controls are created by perturbing a single component of an otherwise correct synthesis route. Thus, retrieving the correct paper alone is insufficient; the system must recover the reported experimental conditions precisely enough to detect the altered detail. T-MOF-2 therefore measures the reliability of literature-based synthesis-information retrieval and verification.

\textbf{T-MOF-3: Evidence-Sufficiency Verification.}
T-MOF-3 evaluates whether a model can judge scientifically contested MOF hypotheses by reconciling evidence across the literature rather than relying on a single source. Each item presents a proposition for which different studies may provide supporting, contradicting, or condition-dependent evidence. The model must determine whether the hypothesis is \textit{Yes}, when the available literature consistently supports it; \textit{No}, when the literature provides sufficient evidence to refute it; or \textit{Uncertain}, when the conclusion depends on experimental conditions, conflicting reports or literature, or evidence that is insufficient to justify a unique verdict. The dataset contains 117 one-sentence propositions spanning structure--property relationships, synthesis and activation protocols, characterization limits, and comparisons across materials or experimental conditions. For example, high-pressure in situ single-crystal X-ray diffraction on a flexible Cd--MOF directly resolved the reversible closed- and open-pore structures associated with CO$_2$ adsorption, whereas the CO$_2$ binding sites were assigned using combined single-crystal and powder diffraction evidence. A claim that attributes both observations to a single measurement modality is therefore not fully supported by the available evidence and should be treated as \textit{Uncertain}. T-MOF-3 thus tests whether a model can distinguish well-supported conclusions from claims whose validity is conditional, disputed, or underdetermined by the literature.

\textit{Controlled evaluation for T-MOF-1--3.}
We evaluate T-MOF-1--3 under three controlled evidence-access settings: \textbf{closed-book}, where the LLM relies only on its parametric knowledge; \textbf{retrieval-enabled}, where external evidence is retrieved through RAG~\cite{lewis2020rag}; and \textbf{oracle-evidence}, where the evidence required for verification is directly provided. Across these settings, the model predicts one of three verdicts: \textit{Yes}, \textit{No}, or \textit{Uncertain}. This controlled design allows us to examine how performance changes as evidence access is progressively improved, thereby localizing whether a failure is associated with limited knowledge access, evidence retrieval, or reasoning over the available evidence.

\textbf{T-MOF-4: MLIP-based Computational Verification.}
T-MOF-4 targets hypotheses that cannot be resolved through structural or literature evidence alone and instead require an explicit computation. For each item, the reference outcome is derived from an MLIP calculation executed under a fixed specification. Unlike the three-condition T-MOF-1--3 protocol, T-MOF-4 compares no-tool Closed-book (Track 1) with the computational harness (Track 4); RAG and Oracle are inapplicable. The resulting verdict is \textit{Yes} when the calculation yields a valid result, \textit{Re-run} when the calculation does not converge, and \textit{Human-review} when the requested quantity falls outside what the calculator can determine. T-MOF-4 therefore evaluates whether the Co-Scientist can obtain and correctly use computational evidence to settle a hypothesis.

\textit{Evidence grounding and benchmark controls.}
For each task, we explicitly record the evidence required to justify the reference verdict: a CSD refcode for T-MOF-1, an exact literature record with DOI for T-MOF-2, the decisive condition specified by our MOF expert for T-MOF-3, and a fixed MLIP calculation specification for T-MOF-4. This allows us to evaluate not only whether the final verdict is correct, but also whether it is supported by the appropriate evidence.
For T-MOF-1--3, a fixed no-tool GPT-5.4 baseline characterizes answerability; both lower-answerability and lookup strata are retained (Appendix~\ref{sec:implementation-details}). All items are constructed by trained annotators in MOF using the CSD and primary literature under conventions ratified by a MOF expert holding a Ph.D. in the field; the expert also authors the T-MOF-3 reasoning items and verifies the decoy construction.

\begin{table*}[t]
  \centering
  \scriptsize
  \setlength{\tabcolsep}{3pt}
  \renewcommand{\arraystretch}{0.72}

  \caption{Macro-F1 across T-MOF-1--3 under Closed-book, Retrieval-enabled, and Oracle-evidence settings, and T-MOF-4 comparing Closed-book with the computational harness, with \textsc{MOF-Verify} (Ours) for comparison. (Note: T-MOF-1 and T-MOF-4 in \textsc{MOF-Verify} are LLM-independent rule-based tasks, yielding identical Macro-F1 across models (\texttt{78.31} and \texttt{56.97}, respectively), which we report for completeness. The best result among Closed, RAG, Oracle, and Ours is shown in \textbf{bold}.)}
  \label{tab:main-results}

  \begin{tabular}{
    l
    l
    c
    >{\centering\arraybackslash}p{1.25cm}
    >{\centering\arraybackslash}p{1.25cm}
    >{\centering\arraybackslash}p{1.25cm}|
    >{\centering\arraybackslash}p{1.25cm}|
    >{\centering\arraybackslash}p{1.25cm}|
    >{\centering\arraybackslash}p{1.45cm}
  }
    \toprule
    \textbf{Family}
    & \textbf{Model}
    & \textbf{Dataset}
    & \textbf{Closed}
    & \textbf{RAG}
    & \textbf{Oracle}
    & $\boldsymbol{\Delta}_{\textbf{access}}$
    & $\boldsymbol{\Delta}_{\textbf{retrieval}}$
    & \textbf{Ours} \\
    \midrule

    \multirow{16}{*}{OpenAI}
    & \multirow{4}{*}{gpt-5.4}
    & T-MOF-1 & 11.09 & 55.99 & \textbf{82.86} & 44.89 & 26.87 & 78.31 \\
    & & T-MOF-2 & 24.47 & 51.76 & \textbf{60.96} & 27.29 & 9.20 & 50.00 \\
    & & T-MOF-3 & 58.09 & 43.19 & \textbf{64.52} & -14.90 & 21.33 & 54.40 \\
    & & T-MOF-4 & 16.67 & -- & -- & -- & -- & 56.97 \\
    \cmidrule(lr){2-9}

    & \multirow{4}{*}{gpt-5.6-terra}
    & T-MOF-1 & 19.39 & 62.39 & \textbf{80.64} & 43.00 & 18.25 & 78.31 \\
    & & T-MOF-2 & 39.70 & 53.19 & \textbf{62.39} & 13.50 & 9.20 & 50.78 \\
    & & T-MOF-3 & 61.26 & 44.82 & 56.34 & -16.44 & 11.52 & \textbf{58.07} \\
    & & T-MOF-4 & 0.00 & -- & -- & -- & -- & 56.97 \\
    \cmidrule(lr){2-9}

    & \multirow{4}{*}{gpt-5.6-sol}
    & T-MOF-1 & 14.11 & 62.94 & \textbf{85.82} & 48.83 & 22.88 & 78.31 \\
    & & T-MOF-2 & 23.44 & 53.35 & \textbf{62.64} & 29.92 & 9.29 & 54.62 \\
    & & T-MOF-3 & 57.49 & 49.26 & 61.13 & -8.23 & 11.87 & \textbf{63.75} \\
    & & T-MOF-4 & 0.00 & -- & -- & -- & -- & 56.97 \\
    \cmidrule(lr){2-9}

    & \multirow{4}{*}{gpt-4o}
    & T-MOF-1 & 3.12 & 59.52 & 74.61 & 56.41 & 15.09 & \textbf{78.31} \\
    & & T-MOF-2 & 4.34 & 55.12 & \textbf{67.55} & 50.78 & 12.43 & 46.61 \\
    & & T-MOF-3 & 28.20 & 54.29 & \textbf{67.88} & 26.09 & 13.59 & 48.69 \\
    & & T-MOF-4 & 14.81 & -- & -- & -- & -- & 56.97 \\

    \midrule

    \multirow{12}{*}{Anthropic}
    & \multirow{4}{*}{claude-opus-4.8}
    & T-MOF-1 & 7.45 & 63.26 & 76.97 & 55.81 & 13.71 & \textbf{78.31} \\
    & & T-MOF-2 & 11.24 & 51.17 & \textbf{58.81} & 39.93 & 7.64 & 57.03 \\
    & & T-MOF-3 & 66.72 & 50.53 & 61.63 & -16.19 & 11.10 & \textbf{70.56} \\
    & & T-MOF-4 & 0.00 & -- & -- & -- & -- & 56.97 \\
    \cmidrule(lr){2-9}

    & \multirow{4}{*}{claude-opus-5}
    & T-MOF-1 & 16.14 & 63.71 & \textbf{79.90} & 47.57 & 16.19 & 78.31 \\
    & & T-MOF-2 & 16.99 & 53.64 & \textbf{63.19} & 36.65 & 9.55 & 56.01 \\
    & & T-MOF-3 & 62.50 & 52.72 & 58.71 & -9.78 & 5.99 & \textbf{65.99} \\
    & & T-MOF-4 & 0.00 & -- & -- & -- & -- & 56.97 \\
    \cmidrule(lr){2-9}

    & \multirow{4}{*}{claude-sonnet-5}
    & T-MOF-1 & 6.79 & 60.93 & 68.06 & 54.14 & 7.13 & \textbf{78.31} \\
    & & T-MOF-2 & 12.74 & 49.02 & 53.18 & 36.27 & 4.16 & \textbf{55.01} \\
    & & T-MOF-3 & 55.98 & 52.78 & 60.29 & -3.20 & 7.51 & \textbf{61.80} \\
    & & T-MOF-4 & 14.81 & -- & -- & -- & -- & 56.97 \\

    \midrule

    \multirow{4}{*}{Google}
    & \multirow{4}{*}{gemini-3.7-flash}
    & T-MOF-1 & 23.49 & 58.44 & \textbf{87.05} & 34.95 & 28.61 & 78.31 \\
    & & T-MOF-2 & 32.04 & 51.83 & \textbf{64.15} & 19.79 & 12.32 & 51.23 \\
    & & T-MOF-3 & 61.57 & 46.43 & \textbf{62.62} & -15.14 & 16.19 & 57.55 \\
    & & T-MOF-4 & 11.11 & -- & -- & -- & -- & 56.97 \\

    \midrule

    \multirow{4}{*}{Qwen}
    & \multirow{4}{*}{qwen3.7-max}
    & T-MOF-1 & 18.68 & 62.47 & 77.94 & 43.79 & 15.47 & \textbf{78.31} \\
    & & T-MOF-2 & 31.17 & 52.06 & \textbf{59.13} & 20.89 & 7.07 & 47.61 \\
    & & T-MOF-3 & 53.44 & 47.85 & \textbf{57.70} & -5.58 & 9.85 & 50.35 \\
    & & T-MOF-4 & 0.00 & -- & -- & -- & -- & 56.97 \\

    \midrule

    \multirow{4}{*}{DeepSeek}
    & \multirow{4}{*}{deepseek-v4-pro}
    & T-MOF-1 & 21.61 & 70.01 & \textbf{80.76} & 48.40 & 10.75 & 78.31 \\
    & & T-MOF-2 & 29.84 & 57.19 & \textbf{60.68} & 27.35 & 3.49 & 47.11 \\
    & & T-MOF-3 & 57.60 & 49.95 & \textbf{59.55} & -7.65 & 9.60 & 54.49 \\
    & & T-MOF-4 & 0.00 & -- & -- & -- & -- & 56.97 \\

    \midrule
    \multicolumn{3}{l}{\textbf{Average over T-MOF-1--3}}
    & 31.02 & 54.66 & 67.26 & 23.64 & 12.59 & 62.82 \\

    \bottomrule
  \end{tabular}

  \renewcommand{\arraystretch}{1.0}
\end{table*}

\subsection{Diagnosing MOF Verification Bottlenecks with T-MOF-1--4}
\label{sec:diagnosis}

We diagnose MOF verification failures along two complementary axes. 
First, controlled evidence-access settings identify whether performance is limited by parametric knowledge, retrieval, or evidence-conditioned reasoning. 
Second, incorrect predictions are analyzed using the task-specific criteria and rule-based or LLM-judge procedures described below.

\paragraph{(1) Diagnostic protocol and overall performance.}
T-MOF-1--3 use three labels---\textit{Yes}, \textit{No}, and \textit{Uncertain}---with unequal class frequencies. 
We therefore use three-class Macro-F1 as the primary metric, which weights each verdict equally and avoids majority-class performance dominating the evaluation, as can occur with overall accuracy:
\begin{equation}
F1_{\mathrm{macro}}
=
\frac{1}{3}
\sum_{c \in \{\textit{Yes},\textit{No},\textit{Uncertain}\}}
\frac{2P_cR_c}{P_c+R_c},
\label{eq:macro-f1}
\end{equation}
where $P_c$ and $R_c$ denote class-wise precision and recall.
Table~\ref{tab:main-results} reports Macro-F1 across models and controlled evidence-access settings; the corresponding class-wise F1 results are provided in Appendix~\ref{sec:tmof-complete}, Tables~\ref{tab:classwise-results1} and~\ref{tab:classwise-results}.

Rather than treating Closed-book, RAG, and Oracle as independent baselines, we use their performance gaps as diagnostic signals:
\begin{equation}
\Delta_{\mathrm{access}}
=
F1_{\mathrm{RAG}}-F1_{\mathrm{Closed}},
\qquad
\Delta_{\mathrm{retrieval}}
=
F1_{\mathrm{Oracle}}-F1_{\mathrm{RAG}}.
\label{eq:diagnostic-gaps}
\end{equation}
A large positive $\Delta_{\mathrm{access}}$ indicates that external evidence compensates for limitations in parametric knowledge, whereas a large positive $\Delta_{\mathrm{retrieval}}$ indicates that relevant evidence is useful once provided but is not reliably retrieved or grounded by RAG. $F1_{\mathrm{Oracle}}$ measures performance when the model is given golden evidence sufficient to derive the ground-truth answer, serving as an evidence-access reference.

\textbf{Diagnostic analysis.} Across T-MOF-1--3, RAG improves Macro-F1 over Closed-book by 23.64 points on average, while Oracle provides a further 12.59-point gain, confirming that both evidence access and retrieval quality are major bottlenecks. The pattern, however, differs substantially by task. T-MOF-1 shows the largest gains from both RAG and Oracle ($\Delta_{\mathrm{access}}=47.78$, $\Delta_{\mathrm{retrieval}}=17.50$ on average), indicating that structural verification is dominated by missing parametric knowledge and unreliable structure retrieval. T-MOF-2 similarly benefits from evidence access ($\Delta_{\mathrm{access}}=30.24$), but its smaller Oracle gain ($\Delta_{\mathrm{retrieval}}=8.44$) suggests that synthesis verification is less retrieval-limited once the relevant literature is available. In contrast, T-MOF-3 exhibits a negative average $\Delta_{\mathrm{access}}$ of $-7.10$: naive RAG often degrades performance when retrieved evidence is conflicting or condition-dependent. Oracle evidence recovers 11.86 points on average, yet reaches only 61.04 Macro-F1, indicating that evidence-sufficiency reasoning remains difficult even when decisive evidence is provided. Together, these results localize distinct bottlenecks in structural grounding, synthesis-evidence acquisition, and evidence-conditioned reasoning rather than a single generic retrieval failure.

\paragraph{(2) Task-specific failure attribution.}
Aggregate performance gaps indicate \emph{where} evidence access matters, but do not reveal the concrete cause of an incorrect verdict.
We therefore further analyze the rationales accompanying incorrect predictions using task-specific attribution procedures.
For T-MOF-1 and T-MOF-2, we apply rule-based screening algorithms to determine whether each incorrect prediction satisfies predefined structural- or synthesis-related failure criteria based on its rationale.
For T-MOF-3, where evidence insufficiency depends more strongly on contextual interpretation, we use GPT-5.6 Sol as an LLM judge to classify each incorrect rationale into the predefined evidence-insufficiency categories.
In all cases, a failure category is assigned only when the rationale explicitly supports the corresponding diagnostic cause, rather than simply because the final verdict differs from the ground-truth label.
Because multiple failure categories may apply to the same prediction, the categories are non-mutually exclusive, and each diagnostic rate is computed over the union of relevant errors.

\textbf{T-MOF-1: Structural identity and grounding.}
We distinguish \emph{wrong material identity} from \emph{missing or untrusted deposited structure}.
Let $\mathcal{I}_1$ denote the set of incorrect T-MOF-1 predictions, and let $\mathcal{E}_{\mathrm{id}}$ and $\mathcal{E}_{\mathrm{str}}$ denote identity and structural-evidence errors, respectively.
We define
$R_{\mathrm{struct}}
=
|\mathcal{I}_1 \cap (\mathcal{E}_{\mathrm{id}}\cup\mathcal{E}_{\mathrm{str}})|
/|\mathcal{I}_1|$,
which measures the fraction of incorrect predictions attributable to structural identity or grounding failures.

\textbf{T-MOF-2: Synthesis-condition retrieval.}
We distinguish failure to retrieve the required synthesis evidence from failure to correctly read or verify a retrieved condition.
Let $\mathcal{I}_2$ denote the set of incorrect T-MOF-2 predictions, and let $\mathcal{E}_{\mathrm{miss}}$ and $\mathcal{E}_{\mathrm{misread}}$ denote missing-evidence and misreading errors, respectively.
We define
$R_{\mathrm{synth}}
=
|\mathcal{I}_2 \cap (\mathcal{E}_{\mathrm{miss}}\cup\mathcal{E}_{\mathrm{misread}})|
/|\mathcal{I}_2|$.
This distinction is important because T-MOF-2 controls perturb only one component of an otherwise correct synthesis route; overlooking a single temperature, solvent, time, modulator, or precursor can therefore produce a plausible but incorrect verdict.

\textbf{T-MOF-3: Evidence insufficiency.}
T-MOF-3 focuses on cases in which the literature is conflicting, a decisive condition is missing, or a claim is valid only under restricted conditions.
Let $\mathcal{I}_3$ denote the set of incorrect T-MOF-3 predictions, and let $\mathcal{E}_{\mathrm{insuff}}$ denote the set of these evidence-insufficiency failures.
We define
$R_{\mathrm{evidence}}
=
|\mathcal{I}_3 \cap \mathcal{E}_{\mathrm{insuff}}|
/|\mathcal{I}_3|$.
Because the central failure is an unsupported commitment when the evidence does not justify a unique conclusion, we additionally measure
$R_{\mathrm{over}}
=
N(y=\textit{Uncertain},\hat{y}\in\{\textit{Yes},\textit{No}\})
/
N(y=\textit{Uncertain})$.
This measures how often the model converts an evidence-insufficient case into an unsupported binary verdict.

\textbf{Diagnostic analysis.}
Table~\ref{tab:diagnostic-errors} shows that the diagnosed failure modes account for a substantial fraction of errors across models and evidence-access settings. Averaged over all models and Closed, RAG, and Oracle, $R_{\mathrm{struct}}$, $R_{\mathrm{synth}}$, and $R_{\mathrm{evidence}}$ are 94.87\%, 93.77\%, and 64.66\%, respectively, while $R_{\mathrm{over}}$ shows that 81.18\% of gold-\textit{Uncertain} cases are incorrectly committed to \textit{Yes} or \textit{No}. These rates summarize the task-specific attributions; the LLM-based attributions have not been independently validated. The 81.18\% overcommitment rate concerns baseline gold-\textit{Uncertain} cases, not the harness ES error rate. Importantly, the average diagnostic rates remain high under Closed, RAG, and Oracle (82.47\%, 81.40\%, and 87.00\%, respectively), indicating that improved evidence access alone does not eliminate these task-specific bottlenecks. This motivates \textsc{MOF-Verify} to explicitly target structural grounding, synthesis-condition retrieval, and evidence sufficiency with dedicated verification modules.

\begin{table*}[t]
  \centering
  \scriptsize
  \setlength{\tabcolsep}{4pt}
  \renewcommand{\arraystretch}{0.72}

  \caption{Task-specific failure rates (unit: \%). Structural, synthesis, and evidence rates condition on incorrect predictions; overcommitment conditions on gold-\textit{Uncertain} cases. Average denotes the unweighted macro-average across Closed, RAG, and Oracle. Cases quarantined because of unresolved annotation or gold-integrity issues are excluded from the corresponding denominators.}
  \label{tab:diagnostic-errors}

  \begin{tabular}{
    l
    l
    c|
    c
    >{\centering\arraybackslash}p{1.25cm}
    >{\centering\arraybackslash}p{1.25cm}
    >{\centering\arraybackslash}p{1.25cm}|
    >{\centering\arraybackslash}p{1.25cm}
  }
    \toprule
    \textbf{Family}
    & \textbf{Model}
    & \textbf{Dataset}
    & \textbf{Metric}
    & \textbf{Closed}
    & \textbf{RAG}
    & \textbf{Oracle}
    & \textbf{Average} \\
    \midrule

    \multirow{16}{*}{OpenAI}
    & \multirow{4}{*}{gpt-5.4}
    & T-MOF-1 & $R_{\mathrm{struct}}$ & 98.18 & 98.99 & 92.31 & 96.49 \\
    & & T-MOF-2 & $R_{\mathrm{synth}}$ & 99.26 & 93.18 & 91.03 & 94.49 \\
    & & \multirow{2}{*}{T-MOF-3} & $R_{\mathrm{evidence}}$ & 60.00 & 57.69 & 57.14 & 58.28 \\
    & & & $R_{\mathrm{over}}$ & 88.89 & 72.22 & 88.24 & 83.12 \\
    \cmidrule(lr){2-8}

    & \multirow{4}{*}{gpt-5.6-terra}
    & T-MOF-1 & $R_{\mathrm{struct}}$ & 97.47 & 100.00 & 76.92 & 91.46 \\
    & & T-MOF-2 & $R_{\mathrm{synth}}$ & 97.64 & 91.42 & 88.78 & 92.61 \\
    & & \multirow{2}{*}{T-MOF-3} & $R_{\mathrm{evidence}}$ & 64.00 & 68.00 & 83.33 & 71.78 \\
    & & & $R_{\mathrm{over}}$ & 88.89 & 83.33 & 100.00 & 90.74 \\
    \cmidrule(lr){2-8}

    & \multirow{4}{*}{gpt-5.6-sol}
    & T-MOF-1 & $R_{\mathrm{struct}}$ & 97.76 & 100.00 & 85.71 & 94.49 \\
    & & T-MOF-2 & $R_{\mathrm{synth}}$ & 99.14 & 93.79 & 87.56 & 93.50 \\
    & & \multirow{2}{*}{T-MOF-3} & $R_{\mathrm{evidence}}$ & 67.86 & 61.54 & 85.71 & 71.70 \\
    & & & $R_{\mathrm{over}}$ & 94.44 & 72.22 & 94.12 & 86.93 \\
    \cmidrule(lr){2-8}

    & \multirow{4}{*}{gpt-4o}
    & T-MOF-1 & $R_{\mathrm{struct}}$ & 99.63 & 100.00 & 88.89 & 96.17 \\
    & & T-MOF-2 & $R_{\mathrm{synth}}$ & 99.88 & 96.34 & 87.42 & 94.55 \\
    & & \multirow{2}{*}{T-MOF-3} & $R_{\mathrm{evidence}}$ & 21.69 & 66.67 & 80.95 & 56.44 \\
    & & & $R_{\mathrm{over}}$ & 27.78 & 83.33 & 76.47 & 62.53 \\

    \midrule

    \multirow{12}{*}{Anthropic}
    & \multirow{4}{*}{claude-opus-4.8}
    & T-MOF-1 & $R_{\mathrm{struct}}$ & 98.43 & 100.00 & 78.57 & 92.33 \\
    & & T-MOF-2 & $R_{\mathrm{synth}}$ & 99.37 & 92.74 & 89.80 & 93.97 \\
    & & \multirow{2}{*}{T-MOF-3} & $R_{\mathrm{evidence}}$ & 51.85 & 48.15 & 85.00 & 61.67 \\
    & & & $R_{\mathrm{over}}$ & 61.11 & 72.22 & 94.12 & 75.82 \\
    \cmidrule(lr){2-8}

    & \multirow{4}{*}{claude-opus-5}
    & T-MOF-1 & $R_{\mathrm{struct}}$ & 98.17 & 100.00 & 68.75 & 88.97 \\
    & & T-MOF-2 & $R_{\mathrm{synth}}$ & 98.79 & 92.59 & 83.96 & 91.78 \\
    & & \multirow{2}{*}{T-MOF-3} & $R_{\mathrm{evidence}}$ & 60.71 & 45.83 & 80.95 & 62.50 \\
    & & & $R_{\mathrm{over}}$ & 77.78 & 66.67 & 100.00 & 81.48 \\
    \cmidrule(lr){2-8}

    & \multirow{4}{*}{claude-sonnet-5}
    & T-MOF-1 & $R_{\mathrm{struct}}$ & 98.46 & 100.00 & 88.24 & 95.57 \\
    & & T-MOF-2 & $R_{\mathrm{synth}}$ & 99.74 & 94.44 & 88.97 & 94.39 \\
    & & \multirow{2}{*}{T-MOF-3} & $R_{\mathrm{evidence}}$ & 52.63 & 57.14 & 81.82 & 63.86 \\
    & & & $R_{\mathrm{over}}$ & 77.78 & 61.11 & 94.12 & 77.67 \\

    \midrule

    \multirow{4}{*}{Google}
    & \multirow{4}{*}{gemini-3.7-flash}
    & T-MOF-1 & $R_{\mathrm{struct}}$ & 97.92 & 100.00 & 100.00 & 99.31 \\
    & & T-MOF-2 & $R_{\mathrm{synth}}$ & 99.15 & 95.54 & 85.42 & 93.37 \\
    & & \multirow{2}{*}{T-MOF-3} & $R_{\mathrm{evidence}}$ & 71.43 & 55.17 & 75.00 & 67.20 \\
    & & & $R_{\mathrm{over}}$ & 83.33 & 66.67 & 94.12 & 81.37 \\

    \midrule

    \multirow{4}{*}{Qwen}
    & \multirow{4}{*}{qwen3.7-max}
    & T-MOF-1 & $R_{\mathrm{struct}}$ & 97.65 & 100.00 & 91.67 & 96.44 \\
    & & T-MOF-2 & $R_{\mathrm{synth}}$ & 99.15 & 93.97 & 91.38 & 94.83 \\
    & & \multirow{2}{*}{T-MOF-3} & $R_{\mathrm{evidence}}$ & 63.64 & 70.37 & 86.36 & 73.46 \\
    & & & $R_{\mathrm{over}}$ & 94.44 & 83.33 & 100.00 & 92.59 \\

    \midrule

    \multirow{4}{*}{DeepSeek}
    & \multirow{4}{*}{deepseek-v4-pro}
    & T-MOF-1 & $R_{\mathrm{struct}}$ & 96.43 & 100.00 & 96.00 & 97.48 \\
    & & T-MOF-2 & $R_{\mathrm{synth}}$ & 98.87 & 95.25 & 88.67 & 94.27 \\
    & & \multirow{2}{*}{T-MOF-3} & $R_{\mathrm{evidence}}$ & 41.67 & 59.26 & 78.26 & 59.73 \\
    & & & $R_{\mathrm{over}}$ & 77.78 & 66.67 & 94.12 & 79.52 \\

    \midrule
    \multicolumn{4}{l}{\textbf{Average}}
    & 82.47
    & 81.40
    & 87.00
    & \\

    \bottomrule
  \end{tabular}

  \renewcommand{\arraystretch}{1.0}
\end{table*}

\section{\textsc{MOF-Verify}: A Failure-Aware Agentic Harness}

We introduce \textsc{MOF-Verify}, an agentic verification framework that operates with a frozen base LLM and requires no additional training or fine-tuning. Failure classes and the harness component that addresses each are summarized in Table~\ref{tab:failure-map} in Appendix~\ref{sec:failure}.

\begin{figure*}[t]
    \centering
\includegraphics[width=0.9\textwidth]{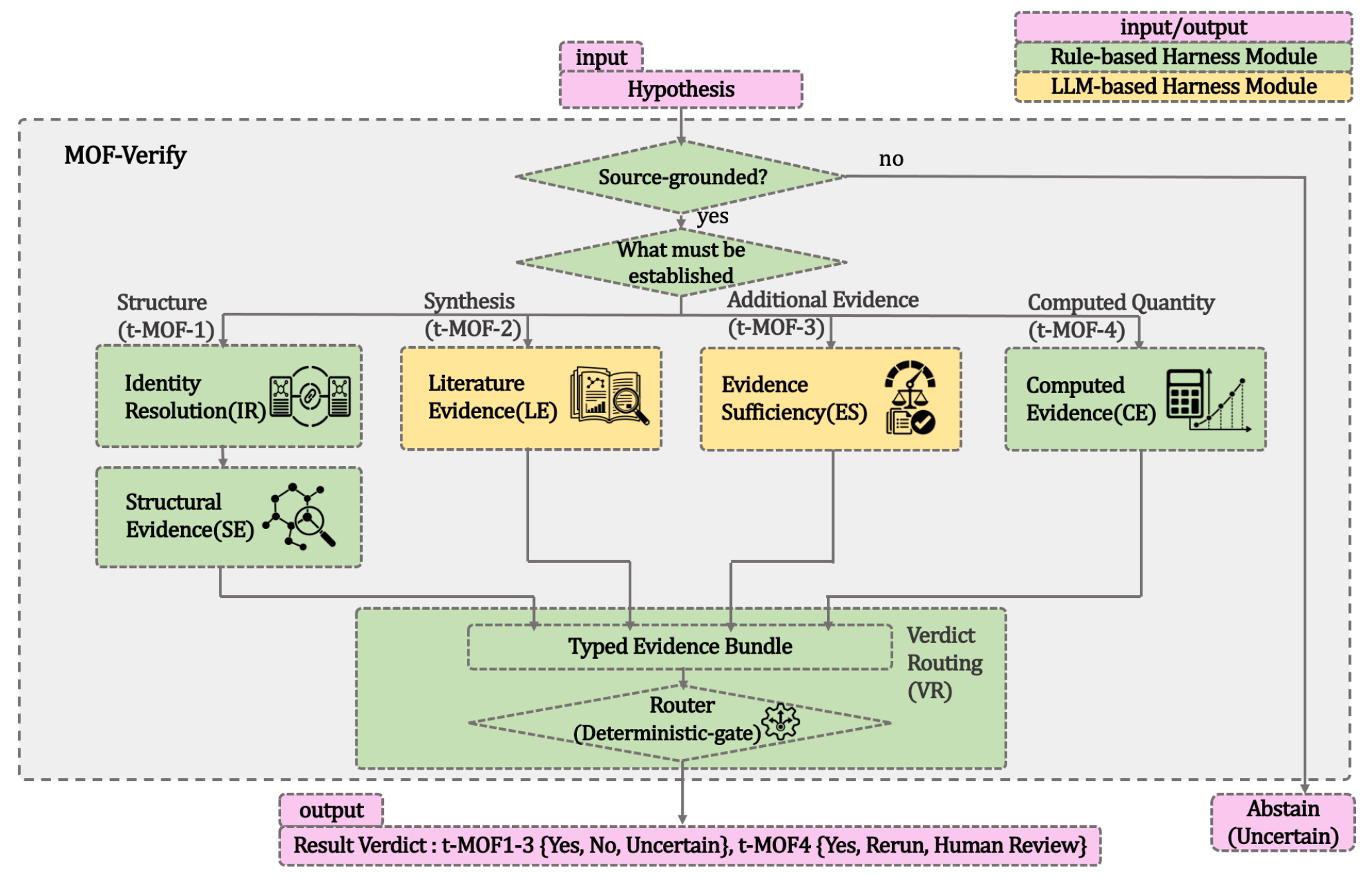}
    \caption{\textbf{Workflow of \textsc{MOF-Verify}.} An input hypothesis uses the task-relevant IR, SE, LE, ES, or CE modules, whose provenance-aware outputs are integrated by a deterministic verdict router to produce the final output verdict or abstention.}
    \label{fig:mof-verify}
\end{figure*}

\subsection{Workflow}
Figure~\ref{fig:mof-verify} shows the system interfaces; the task selects the relevant path rather than invoking every module. Structural questions use IR and SE, synthesis questions use LE, evidence-sufficiency questions use ES, and computation-dependent questions use CE. Evidence-producing modules return typed outputs with sources, provenance, and content hashes. VR uses these outputs to assign the final label. Appendix~\ref{sec:hyptrace} gives a worked trace.

\textbf{Identity Resolution.}
\textbf{IR} maps a material mention to a canonical identity and a deposited-structure anchor, returning a resolution status rather than a name alone. It matches the mention against a curated and versioned table of MOF names and aliases, ensuring that changes to the resolver's knowledge are explicitly recorded. When normalization and lookup leave a mention ambiguous or unmatched, the mention is returned as \textit{unresolved}. 

\textbf{Structural Evidence.}
\textbf{SE} maps the canonical material identity from \textbf{IR} to a trusted, provenance-carrying CIF through deterministic lookup. The crystallographic file must match the digest, a content hash recorded when the structure was vetted. Candidates are ranked experimental-first using CSD records, so that a computed structure, such as a QMOF entry, is not returned as the trusted deposited structure. \textbf{IR} and \textbf{SE} make no model call. 

\textbf{Literature Evidence.}
Expert-supplied papers are acquired in advance and stored with their DOIs and digests. Missing or altered documents are rejected. LE reads the paper's main text and supporting information to check the claimed synthesis conditions and collect relevant literature evidence. Reported conditions are retained with supporting source passages. Conflicting reported values are surfaced, not silently selected, and an unavailable DOI is reported as such rather than answered from memory.

\textbf{Evidence Sufficiency.} \textbf{ES} determines whether the available evidence is sufficient to adjudicate a given hypothesis. Rather than treating all retrieved evidence as equally reliable, \textbf{ES} first evaluates evidence quality based on traceability, corroboration, and material grounding, and provenance quality based on the completeness and trustworthiness of the underlying sources. It then checks whether the evidence contains conflicting findings or lacks conditions and information required to resolve the claim. The resulting quality scores, conflict records, and evidence-gap indicators are passed to \textbf{VR}, which determines the final verdict and abstains when the evidence is insufficient for a reliable decision.

\textbf{Verdict Routing.}
\textbf{VR} is a deterministic router that determines whether the assembled evidence is sufficient to support a scientific verdict. It evaluates provenance completeness and evidence quality before applying a fixed sequence of decision gates. A decisive \textit{Yes} or \textit{No} requires that the evidence source is recognized, provenance satisfies the required checks, trusted sources do not contain unresolved contradictions, and any computational evidence used in the decision is valid. When these conditions are not met, \textbf{VR} returns an abstaining or \textit{Uncertain} rather than forcing a binary conclusion. The paper-reading prompt supplies an LLM-proposed verdict with a quote and rationale; VR applies the evidence and validation checks before assigning the final label. 

\textbf{Computational Evidence.}
\textbf{CE} handles computational hypotheses by converting the requested quantity into an executable specification that fixes the structure, model, precision, device, seed, and convergence criteria. It then executes the specification using the MLIP model SevenNet~\citep{kim2025data,park2024scalable} and returns either the computed property together with its convergence record or a typed refusal indicating that the calculation did not converge or that the requested task is unsupported. The computational output is passed to \textbf{VR}, which determines the T-MOF-4 verdict: \textit{Yes} when a valid result is obtained, \textit{Re-run} when the calculation does not converge, and \textit{Human-review} when the requested quantity falls outside what the calculator can determine.

\subsection{Performance Analysis}

As shown in Table~\ref{tab:main-results}, \textsc{MOF-Verify} achieves an average Macro-F1 of 62.82 across T-MOF-1--3, improving over Closed-book and RAG by 31.80 and 8.16 points, respectively, with an Oracle evidence-access reference of 67.26. The gains are strongly task-dependent and align with the diagnosed bottlenecks. On T-MOF-1, the deterministic IR+SE pipeline reaches 78.31 Macro-F1, improving over RAG by 16.34 points on average and coming within 1.15 points of Oracle, indicating that explicit identity resolution and trusted-structure grounding largely close the structural-evidence gap. On T-MOF-3, \textsc{MOF-Verify} improves over RAG by 9.38 points on average and even exceeds Oracle for several strong backbones, showing higher full-system scores under these settings without isolating the contributions of ES and VR. T-MOF-2 remains more challenging: performance is 1.23 points below RAG on average and below Oracle, indicating that precise synthesis-condition extraction remains a residual bottleneck. Finally, on T-MOF-4, CE raises average performance from 5.74 under direct LLM inference to 56.97, in the eight-item computational task; this comparison changes tool access and does not isolate orchestration from tool availability. Overall, these results support the failure-aware design of \textsc{MOF-Verify}: the combined harness improves verification on this benchmark, while also revealing where further improvement is still needed.

\paragraph{Scope and limitations.}
The computational task contains eight items. The evaluation does not include resource-matched component ablations, held-out testing, or repeated-run confidence intervals. Independent validation of the LLM-based error attribution, evidence-sufficiency criteria, and verdict-routing rules remains necessary.

\section{Conclusion}

MOF hypothesis verification is a system problem: a plausible answer can still refer to the wrong structure, omit a synthesis condition, overstate conflicting evidence, or skip a necessary calculation. Our four-task benchmark separates these challenges, and \textsc{MOF-Verify} connects them to task-specific evidence acquisition and verdict routing without fine-tuning the base LLM.

The reported gains are strongest for structural grounding and evidence sufficiency, while synthesis verification remains a bottleneck. These differences make the benchmark useful for deciding which part of a verification workflow needs attention, rather than treating more retrieval or more modules as a universal remedy. The provenance-carrying outputs and explicit unresolved states may also help practitioners inspect the evidence behind a verdict before deciding what to verify next.

The present results support this design on a curated benchmark, not general autonomous discovery or independently validated sufficiency judgments. Broader held-out evaluation and resource-matched comparisons are needed to test transfer and isolate component benefits.

\clearpage
\begin{ack}
This research was supported by the National Research Foundation of Korea (NRF) grant funded by the Korea government (MSIT and MOE) (RS-2025-16063688), and the Korea Basic Science Institute (National Research Facilities and Equipment Center) grant funded by the Ministry of Science and ICT (RS-2024-00404602).
\end{ack}

\bibliographystyle{plainnat}
 \bibliography{references}

\newpage

\appendix

\section{Technical appendices and supplementary material}

\subsection{Sampled example hypothesis of T-MOF-1-4}
\label{sec:sample-hypotheses}

\paragraph{(1) T-MOF-1}

\begin{itemize}

\item \textbf{Example-1}

\noindent
\fbox{
\begin{minipage}{0.95\linewidth}
\textbf{Hypothesis:} A resolver proposes CSD refcode OCUNAC as the trusted structure for UiO-66. Should it be accepted?

\vspace{2pt}
\textbf{Answer:} No
\end{minipage}
}

\vspace{3pt}
\textbf{Justification:} OCUNAC is MIL-101, a Cr-based framework with a different topology, whereas UiO-66 is Zr$_6$O$_4$(OH)$_4$(BDC)$_6$ and is anchored by RUBTAK. Therefore, OCUNAC cannot be accepted as the trusted UiO-66 structure.

\item \textbf{Example-2}

\noindent
\fbox{
\begin{minipage}{0.95\linewidth}
\textbf{Hypothesis:} Does the simulated structural model reported for IRMOF-13 represent the same framework structure as the experimentally refined IRMOF-14 structure (CSD refcode EDUVOO)?

\vspace{2pt}
\textbf{Answer:} No
\end{minipage}
}

\vspace{3pt}
\textbf{Justification:} IRMOF-13 is a simulated, doubly interpenetrated trigonal R$\bar{3}$m model, whereas IRMOF-14 (EDUVOO) is an experimentally refined, non-interpenetrated cubic Fm$\bar{3}$m structure. Their shared Zn$_4$O/PDC composition and pcu topology do not make the framework structures identical.

\end{itemize}

\paragraph{(2) T-MOF-2}

\begin{itemize}

\item \textbf{Example-1}

\noindent
\fbox{
\begin{minipage}{0.95\linewidth}
\textbf{Hypothesis:} According to DOI 10.1002/anie.202000278, was the methyl-iodide (CH$_3$I) methylation of Co$_2$(m-DOBDC) performed at 60\,$^{\circ}$C?

\vspace{2pt}
\textbf{Answer:} No
\end{minipage}
}

\vspace{3pt}
\textbf{Justification:} The supporting information reports CH$_3$I treatment at 100\,$^{\circ}$C for 48 h. The 60\,$^{\circ}$C condition belongs to formaldehyde treatment, so the hypothesis swaps the temperatures of two different modifications.

\item \textbf{Example-2}

\noindent
\fbox{
\begin{minipage}{0.95\linewidth}
\textbf{Hypothesis:} In Wang et al.\ (Chem.\ Eur.\ J.\ 2013, DOI 10.1002/chem.201300144), the porous Zn framework designated MOF-1, [Zn$_3$(L)(H$_2$O)$_2$] (L = the hexakis(3-carboxyphenyloxymethylene)benzene hexacarboxylate), was prepared solvothermally. Is the reported route: H$_6$L + Zn(NO$_3$)$_2\cdot$6H$_2$O dissolved in DMF, heated in a Teflon reactor at 78\,$^{\circ}$C for 3 days?

\vspace{2pt}
\textbf{Answer:} Yes
\end{minipage}
}

\vspace{3pt}
\textbf{Justification:} The Experimental Section reports H$_6$L and Zn(NO$_3$)$_2\cdot$6H$_2$O in 8 mL DMF, heated in a 15 mL Teflon reactor at 78\,$^{\circ}$C for 3 days. Every synthesis condition asserted in the hypothesis matches the reported route.

\end{itemize}

\paragraph{(3) T-MOF-3}

\begin{itemize}

\item \textbf{Example-1}

\noindent
\fbox{
\begin{minipage}{0.95\linewidth}
\textbf{Hypothesis:} Do MOF glasses of the same chemical composition have the same glass-transition temperature?

\vspace{2pt}
\textbf{Answer:} No
\end{minipage}
}

\vspace{3pt}
\textbf{Justification:} Qiao et al.\ reported that ZIF-62 glass quenched at 60 MPa had a significantly lower glass-transition temperature than the same-composition glass formed at ambient pressure, so composition alone does not fix $T_g$.

\item \textbf{Example-2}

\noindent
\fbox{
\begin{minipage}{0.95\linewidth}
\textbf{Hypothesis:} A MOF's sharp PXRD Bragg peaks disappear after thermal treatment. Has the material lost its atomic-scale order?

\vspace{2pt}
\textbf{Answer:} Uncertain
\end{minipage}
}

\vspace{3pt}
\textbf{Justification:} Loss of sharp Bragg peaks is compatible with both decomposition and vitrification, but the hypothesis does not specify which occurred. Those routes differ in whether short-range atomic order survives.

\end{itemize}

\paragraph{(4) T-MOF-4}

\begin{itemize}

\item \textbf{Example-1}

\noindent
\fbox{
\begin{minipage}{0.95\linewidth}
\textbf{Hypothesis:} A SevenNet-Omni relaxation of MOF-5 (<=80 steps) targets an over-tight fmax of 1e-6 eV/A. Should its result be accepted as a converged single-phase result?

\vspace{2pt}
\textbf{Answer:} Yes
\end{minipage}
}

\vspace{3pt}
\textbf{Justification:} Yes---the CE module relaxes MOF-5 to \texttt{converged=True} in 70 steps even under the over-tight force threshold $f_{\max}=10^{-6}$~eV/\AA, as this small, high-symmetry cell reaches near-zero forces. The result is therefore genuinely converged. The verdict is byte-deterministic across reruns for a fixed cell: \texttt{spec\_sha256} and the CE-module calculator fingerprint define the stable determinism key, while the raw energy exhibits only $\sim 10^{-6}$~eV CPU-reduction jitter, which does not affect the verdict. Notably, the same $f_{\max}=10^{-6}$~eV/\AA specification does not converge for UiO-66 within 80 steps, showing that the verdict is a deterministic function of the structure rather than the tolerance alone.

\item \textbf{Example-2}

\noindent
\fbox{
\begin{minipage}{0.95\linewidth}
\textbf{Hypothesis:} A SevenNet-Omni relaxation of UiO-66 was run with a budget of \texttt{max\_steps=1} and $f_{\max}=0.05$~eV/\AA. Should its energy be accepted as a converged single-phase result?

\vspace{2pt}
\textbf{Answer:} Re-run
\end{minipage}
}

\vspace{3pt}
\textbf{Justification:} Re-run---CE returns a typed \texttt{non\_converged} refusal: a budget of \texttt{max\_steps=1} is insufficient to relax UiO-66, which requires tens of optimization steps. The resulting one-step energy is therefore not a converged value and should not be accepted. The calculation should be re-run with an adequate step budget. This refusal is deterministic across reruns.

\end{itemize}

\subsection{Complete Results for T-MOF-1--3}
\label{sec:tmof-complete}

Table~\ref{tab:classwise-results1} reports class-wise F1 for T-MOF-1--3 under the closed-book, retrieval-enabled, and oracle-evidence settings. Table~\ref{tab:classwise-results} reports the corresponding results for T-MOF-2--3 with \textsc{MOF-Verify}.

\begin{table*}[!t]
  \centering
  \footnotesize
  \setlength{\tabcolsep}{2.5pt}
  \renewcommand{\arraystretch}{0.72}

  \caption{Class-wise F1 across T-MOF-1--3 under Closed-book, Retrieval-enabled, and Oracle-evidence settings.}
  \label{tab:classwise-results1}

  \begin{tabular}{
    l
    l
    c|
    ccc|
    ccc|
    ccc
  }
    \toprule
    \textbf{Family}
    & \textbf{Model}
    & \textbf{Dataset}
    & \multicolumn{3}{c|}{\textbf{Closed}}
    & \multicolumn{3}{c|}{\textbf{RAG}}
    & \multicolumn{3}{c}{\textbf{Oracle}} \\

    \cmidrule(lr){4-6}
    \cmidrule(lr){7-9}
    \cmidrule(lr){10-12}

    &
    &
    & \textbf{Yes}
    & \textbf{No}
    & \textbf{Uncertain}
    & \textbf{Yes}
    & \textbf{No}
    & \textbf{Uncertain}
    & \textbf{Yes}
    & \textbf{No}
    & \textbf{Uncertain} \\
    \midrule

    \multirow{12}{*}{OpenAI}
    & \multirow{3}{*}{gpt-5.4}
    & T-MOF-1
    & 0.00 & 31.62 & 1.67
    & 66.67 & 85.40 & 15.89
    & 90.70 & 91.23 & 66.67 \\
    & & T-MOF-2
    & 34.98 & 25.10 & 13.33
    & 61.85 & 74.20 & 19.23
    & 73.78 & 81.84 & 27.27 \\
    & & T-MOF-3
    & 76.92 & 80.67 & 16.67
    & 33.33 & 69.57 & 26.67
    & 88.31 & 86.21 & 19.05 \\
    \cmidrule(lr){2-12}

    & \multirow{3}{*}{gpt-5.6-terra}
    & T-MOF-1
    & 0.00 & 58.18 & 0.00
    & 78.79 & 87.88 & 20.51
    & 96.68 & 95.24 & 50.00 \\
    & & T-MOF-2
    & 52.55 & 46.33 & 20.22
    & 69.02 & 76.28 & 14.29
    & 78.98 & 84.49 & 23.70 \\
    & & T-MOF-3
    & 80.43 & 84.30 & 19.05
    & 33.33 & 63.64 & 37.50
    & 83.54 & 85.47 & 0.00 \\
    \cmidrule(lr){2-12}

    & \multirow{3}{*}{gpt-5.6-sol}
    & T-MOF-1
    & 19.16 & 23.18 & 0.00
    & 78.92 & 89.55 & 20.34
    & 95.91 & 94.89 & 66.67 \\
    & & T-MOF-2
    & 32.56 & 25.25 & 12.50
    & 66.97 & 76.92 & 16.17
    & 79.95 & 85.75 & 22.22 \\
    & & T-MOF-3
    & 81.72 & 82.76 & 8.00
    & 33.33 & 70.00 & 44.44
    & 87.18 & 86.21 & 10.00 \\
    \cmidrule(lr){2-12}

    & \multirow{3}{*}{gpt-4o}
    & T-MOF-1
    & 0.00 & 5.63 & 3.72
    & 81.86 & 79.81 & 16.90
    & 97.16 & 98.10 & 28.57 \\
    & & T-MOF-2
    & 0.48 & 0.99 & 11.55
    & 73.90 & 74.29 & 17.18
    & 86.64 & 87.17 & 28.83 \\
    & & T-MOF-3
    & 30.19 & 30.56 & 23.85
    & 33.33 & 75.00 & 54.55
    & 84.34 & 85.98 & 33.33 \\

    \midrule

    \multirow{9}{*}{Anthropic}
    & \multirow{3}{*}{claude-opus-4.8}
    & T-MOF-1
    & 1.42 & 20.12 & 0.83
    & 75.45 & 92.31 & 22.02
    & 90.62 & 90.28 & 50.00 \\
    & & T-MOF-2
    & 3.30 & 19.35 & 11.07
    & 64.00 & 73.85 & 15.66
    & 71.20 & 80.04 & 25.20 \\
    & & T-MOF-3
    & 79.52 & 81.74 & 38.89
    & 44.44 & 57.14 & 50.00
    & 87.18 & 87.18 & 10.53 \\
    \cmidrule(lr){2-12}

    & \multirow{3}{*}{claude-opus-5}
    & T-MOF-1
    & 1.42 & 45.95 & 1.06
    & 75.93 & 93.20 & 22.02
    & 94.81 & 94.89 & 50.00 \\
    & & T-MOF-2
    & 9.91 & 30.02 & 11.04
    & 69.81 & 77.25 & 13.86
    & 81.95 & 85.40 & 22.22 \\
    & & T-MOF-3
    & 81.93 & 82.05 & 23.53
    & 44.44 & 66.67 & 47.06
    & 89.47 & 86.67 & 0.00 \\
    \cmidrule(lr){2-12}

    & \multirow{3}{*}{claude-sonnet-5}
    & T-MOF-1
    & 0.00 & 19.61 & 0.77
    & 68.90 & 93.20 & 20.69
    & 86.29 & 91.81 & 26.09 \\
    & & T-MOF-2
    & 4.65 & 22.13 & 11.46
    & 61.54 & 71.59 & 13.92
    & 67.19 & 76.49 & 15.87 \\
    & & T-MOF-3
    & 72.73 & 75.21 & 20.00
    & 25.00 & 66.67 & 66.67
    & 85.33 & 85.00 & 10.53 \\

    \midrule

    \multirow{3}{*}{Google}
    & \multirow{3}{*}{gemini-3.7-flash}
    & T-MOF-1
    & 61.11 & 7.04 & 2.33
    & 70.42 & 87.39 & 17.52
    & 97.10 & 97.38 & 66.67 \\
    & & T-MOF-2
    & 56.15 & 23.43 & 16.54
    & 65.14 & 74.26 & 16.09
    & 80.43 & 85.85 & 26.15 \\
    & & T-MOF-3
    & 81.63 & 80.00 & 23.08
    & 25.00 & 66.67 & 47.62
    & 90.67 & 86.67 & 10.53 \\

    \midrule

    \multirow{3}{*}{Qwen}
    & \multirow{3}{*}{qwen3.7-max}
    & T-MOF-1
    & 1.40 & 54.64 & 0.00
    & 76.99 & 89.55 & 20.87
    & 91.95 & 91.87 & 50.00 \\
    & & T-MOF-2
    & 30.68 & 45.39 & 17.45
    & 65.55 & 73.97 & 16.67
    & 74.82 & 81.21 & 21.37 \\
    & & T-MOF-3
    & 71.91 & 78.40 & 10.00
    & 33.33 & 72.73 & 37.50
    & 85.71 & 87.39 & 0.00 \\

    \midrule

    \multirow{3}{*}{DeepSeek}
    & \multirow{3}{*}{deepseek-v4-pro}
    & T-MOF-1
    & 0.00 & 64.84 & 0.00
    & 79.49 & 97.22 & 33.33
    & 91.05 & 91.23 & 60.00 \\
    & & T-MOF-2
    & 33.27 & 41.51 & 14.73
    & 64.54 & 75.37 & 31.65
    & 64.84 & 77.19 & 40.00 \\
    & & T-MOF-3
    & 70.00 & 75.20 & 27.59
    & 22.22 & 80.00 & 47.62
    & 82.93 & 85.71 & 10.00 \\

    \bottomrule
  \end{tabular}

  \renewcommand{\arraystretch}{1.0}
\end{table*}
\begin{table*}[!t]
  \centering
  \footnotesize
  \setlength{\tabcolsep}{4pt}
  \renewcommand{\arraystretch}{0.72}

  \caption{Class-wise F1 of \textsc{MOF-Verify} (Ours) on T-MOF-2--3.}
  \label{tab:classwise-results}

  \begin{tabular}{
    l
    l
    c|
    ccc
  }
    \toprule
    \textbf{Family}
    & \textbf{Model}
    & \textbf{Dataset}
    & \multicolumn{3}{c}{\textbf{\textsc{MOF-Verify} (Ours)}} \\
    
    \cmidrule(lr){4-6}

    &
    &
    & \textbf{Yes}
    & \textbf{No}
    & \textbf{Uncertain} \\
    \midrule

    \multirow{8}{*}{OpenAI}
    & \multirow{2}{*}{gpt-5.4}
    & T-MOF-2 & 59.65 & 69.74 & 20.62 \\
    & & T-MOF-3 & 62.07 & 63.04 & 38.10 \\
    \cmidrule(lr){2-6}

    & \multirow{2}{*}{gpt-5.6-terra}
    & T-MOF-2 & 58.35 & 72.60 & 21.39 \\
    & & T-MOF-3 & 70.97 & 63.74 & 39.51 \\
    \cmidrule(lr){2-6}

    & \multirow{2}{*}{gpt-5.6-sol}
    & T-MOF-2 & 63.57 & 75.95 & 24.36 \\
    & & T-MOF-3 & 74.63 & 77.23 & 39.39 \\
    \cmidrule(lr){2-6}

    & \multirow{2}{*}{gpt-4o}
    & T-MOF-2 & 55.63 & 65.67 & 18.53 \\
    & & T-MOF-3 & 52.63 & 56.47 & 36.96 \\

    \midrule

    \multirow{6}{*}{Anthropic}
    & \multirow{2}{*}{claude-opus-4.8}
    & T-MOF-2 & 65.65 & 77.97 & 27.48 \\
    & & T-MOF-3 & 81.16 & 82.24 & 48.28 \\
    \cmidrule(lr){2-6}

    & \multirow{2}{*}{claude-opus-5}
    & T-MOF-2 & 64.93 & 77.64 & 25.44 \\
    & & T-MOF-3 & 78.26 & 78.43 & 41.27 \\
    \cmidrule(lr){2-6}

    & \multirow{2}{*}{claude-sonnet-5}
    & T-MOF-2 & 62.79 & 77.45 & 24.78 \\
    & & T-MOF-3 & 72.73 & 71.58 & 41.10 \\

    \midrule

    \multirow{2}{*}{Google}
    & \multirow{2}{*}{gemini-3.7-flash}
    & T-MOF-2 & 59.03 & 72.28 & 22.39 \\
    & & T-MOF-3 & 68.75 & 68.75 & 35.14 \\

    \midrule

    \multirow{2}{*}{Qwen}
    & \multirow{2}{*}{qwen3.7-max}
    & T-MOF-2 & 54.67 & 67.40 & 20.77 \\
    & & T-MOF-3 & 62.30 & 55.42 & 33.33 \\

    \midrule

    \multirow{2}{*}{DeepSeek}
    & \multirow{2}{*}{deepseek-v4-pro}
    & T-MOF-2 & 56.59 & 65.11 & 19.63 \\
    & & T-MOF-3 & 66.67 & 59.77 & 37.04 \\

    \bottomrule
  \end{tabular}

  \renewcommand{\arraystretch}{1.0}
\end{table*}

\subsection{The Example Hypothesis Traced through Our \textsc{MOF-Verify} Workflow}
\label{sec:hyptrace}

\begin{figure}[H]
  \centering\small
  \begin{tabular}{@{}lp{0.26\linewidth}p{0.45\linewidth}@{}}
    \toprule
    \multicolumn{3}{@{}p{0.97\linewidth}@{}}{\emph{Hypothesis} (T-MOF-2). ``In Cryst.\ Growth Des.\
      2015 (DOI 10.1021/acs.cgd.5b00140), HKUST-1 coating B was prepared. Is the reported route:
      a Cu anode + benzene-1,3,5-tricarboxylic acid in ethanol/H$_2$O, heated at 99\,$^{\circ}$C?''}\\
    \midrule
    Stage & Input & Output (typed, with provenance) \\
    \midrule
    \textbf{LE}
    & the cited DOI
    & document digest verified; evidence from the paper's main text and supporting
      information reports \textbf{49\,$^{\circ}$C}, ethanol/H$_2$O,
      with the supporting source sentence \\
    
    compare
    & claimed vs.\ reported
    & temperature \emph{contradicted}
  ($|99-49| = 50\,^{\circ}\mathrm{C}$, against a $\pm 5\,^{\circ}\mathrm{C}$ tolerance);
  solvent agrees \\
    
    \midrule
    \textbf{VR}
    & the evidence bundle
    & \textbf{No}, citing the paper and the quoted sentence \\
    \bottomrule
  \end{tabular}
  \caption{One hypothesis traced through the harness. When a stage cannot ground its step---for example, because of an unrecognized material, unverifiable file, missing literature record, or insufficient evidence---it emits a typed failure rather than silently passing unsupported information forward. \textbf{VR} then returns the corresponding verdict instead of forcing an unsupported conclusion.}
  \label{fig:harness}
\end{figure}

\subsection{Failure Classes and Corresponding \textsc{MOF-Verify} Components}
\label{sec:failure}
Table~\ref{tab:failure-map} shows pinpointed failure classes and the harness component in \textsc{MOF-VERIFY} to address each.

\begin{table}[ht]
  \centering
  \small
  \caption{Failure classes and the harness component that addresses each.}
  \label{tab:failure-map}
  \begin{tabular}{lll}
    \toprule
    Diagnosed failure & MOF consequence & Component \\
    \midrule
    Identity resolution & wrong material identity & \textbf{IR} \\
    Structure acquisition & missing or untrusted deposited structure & \textbf{SE} \\
    Literature acquisition & missing or misread synthesis conditions & \textbf{LE} \\
    Evidence sufficiency & debatable or conditionally correct claim & \textbf{ES} \\
    Verdict routing & unsupported or over-confident verdict & \textbf{VR} (compare) \\
    Missing computational evidence & unresolved computational claim & \textbf{CE} \\
    \bottomrule
  \end{tabular}
  \vspace{-6pt}
\end{table}

\subsection{Evaluation Prompts and Task-Specific Decision Procedures}
\label{sec:evaluation-prompts}

This section reports the exact prompt templates used in the current evaluation
configuration.  The three diagnostic settings in
Table~\ref{tab:classwise-results1} are denoted \emph{closed-book},
\emph{retrieval-enabled}, and \emph{oracle-evidence}.  They apply to
T-MOF-1--3. T-MOF-4 compares Closed-book (Track 1) with the computational-verification path of \textsc{MOF-Verify} (Track 4); Tracks 2 and 3 are inapplicable.  Braced expressions denote fields populated for
each benchmark item.  No benchmark gold label is included in any model input.
Benchmark stratification, recorded paper-reading settings, retrieval details, and
structured evidence resources are described in Appendix~\ref{sec:implementation-details}.

\subsubsection{Closed-book setting (T-MOF-1--3)}

The same template is used for the structural-grounding, synthesis-condition,
and evidence-sufficiency task families; only \texttt{\{question\}} changes
between items.

\begingroup\scriptsize
\begin{verbatim}
You are answering one materials-chemistry benchmark question in CLOSED-BOOK mode.

Rules:
- Use only your internal model knowledge.
- Do not use external tools, web search, code execution, retrieved documents,
  or databases.
- Return exactly one JSON object and no surrounding Markdown.
- The JSON object must contain exactly these fields:
  - "verdict": one of "Yes", "No", "Uncertain", "Re-run", or "Human-review"
  - "confidence": a number from 0.0 to 1.0
  - "rationale": one concise sentence stating the basis for the verdict
  - "evidence": an empty array, because this is the closed-book condition
- Use the five verdict labels according to their ordinary epistemic meaning.
  If you are not confident, abstain honestly rather than guessing.

Question:
{question}
\end{verbatim}
\endgroup

\subsubsection{Retrieval-enabled setting (T-MOF-1--3)}

For all three task families, the model receives the following system prompt.
The task identifier in the user message specifies whether the item is from
T-MOF-1, T-MOF-2, or T-MOF-3.

\begingroup\scriptsize
\begin{verbatim}
You are answering an EV-1 materials-chemistry benchmark item under the
Plain-RAG condition.

Use only the question and the retrieved text excerpts supplied by the user.
Do not use an oracle paper mapping, benchmark gold, structured harness output,
web search, tools, or unstated outside knowledge. Retrieved excerpts may be
incomplete, noisy, duplicated, or irrelevant. If they do not support a
determinate answer, return Uncertain or Human-review rather than guessing.

Return exactly one JSON object with these fields:
- verdict: one of Yes, No, Uncertain, Re-run, Human-review
- confidence: number from 0 to 1
- rationale: concise evidence-grounded explanation
- evidence: list of objects with kind, ref, resolvable, fulltext_read

For every cited retrieved excerpt:
- kind must be primary_literature
- ref must include its chunk_id exactly as shown in the excerpt header;
  include the DOI when present
- resolvable is true only when a DOI is present, otherwise false
- fulltext_read must be false because only retrieved excerpts were supplied

For a T-MOF-1 structure-identity answer only, when a retrieved excerpt
explicitly names the trusted CSD refcode, you may additionally cite one
db_record object. Its ref must be exactly the refcode, resolvable may be true
only when that exact refcode appears in an excerpt, and fulltext_read must be
false. Do not emit cif or calc_result evidence under the Plain-RAG condition.

Do not claim to have read a complete paper. Output JSON only, without a
Markdown fence or prose.
\end{verbatim}
\endgroup

The corresponding item-level user message is:

\begingroup\scriptsize
\begin{verbatim}
BENCHMARK ITEM: {item_id}
TASK: {T-MOF-1|T-MOF-2|T-MOF-3}

QUESTION
{question}

RETRIEVED TEXT EXCERPTS
[RETRIEVED CHUNK rank={rank} chunk_id={chunk_id}
 source={doi:...|file:...} part={main|si}
 locator=tokens-{start}-{end}]
{chunk_text}

... up to eight retrieved chunks ...

Return the required JSON object now.
\end{verbatim}
\endgroup

\subsubsection{Oracle-evidence setting for structural grounding (T-MOF-1)}

T-MOF-1 requires evidence beyond a source-paper attachment because the claim
may depend on a deposited structure and its identifier.  The oracle packet can
therefore contain source-paper or supporting-information excerpts, licensed
CSD records, and CIF text.

\begingroup\scriptsize
\begin{verbatim}
You are answering one T-MOF-1 materials-chemistry benchmark question in
AUGMENTED ORACLE EVIDENCE mode.

You may use only the question and the evidence packet below. Do not browse,
call tools, execute code, or rely on hidden benchmark labels. The packet can
contain source-paper/SI sections, licensed CSD database records, and CIF text.
A record is included because its refcode appears in the question or because an
independently indexed source paper is mapped to it; inclusion alone is not
evidence that the proposed claim is correct. Compare the proposed refcode,
material identity, formula, cell, bibliographic link, and any other supplied
fields. If the supplied evidence is insufficient or internally inconsistent,
abstain honestly rather than guessing.

CSD/CIF material is licensed for this internal evaluation. In the answer, cite
only DOI/refcode/file identifiers; never reproduce raw CIF coordinates or
licensed record text.

Return exactly one JSON object and no surrounding Markdown. It must contain
exactly:
- "verdict": one of "Yes", "No", "Uncertain", "Re-run", or "Human-review"
- "confidence": a number from 0.0 to 1.0
- "rationale": one concise sentence stating the evidence-based reason
- "evidence": a list of objects with exactly "kind", "ref", "resolvable",
  and "fulltext_read"
  - "kind" is one of "primary_literature", "cif", "calc_result", or
    "db_record"
  - "ref" is a DOI, CSD refcode, or other supplied identifier
  - "resolvable" and "fulltext_read" are booleans

Question:
{question}

Oracle evidence packet:
{evidence_packet}
\end{verbatim}
\endgroup

\subsubsection{Oracle-evidence setting for synthesis-condition and
evidence-sufficiency verification (T-MOF-2--3)}

For T-MOF-2 and T-MOF-3, the oracle evidence consists of the source documents
associated with the benchmark item.  The model receives the following system
prompt.

\begingroup\scriptsize
\begin{verbatim}
You are the oracle-paper arm of a frozen materials-science benchmark.

Answer the question and treat the source documents supplied in the user
message as the primary evidence. Do not use web search, tools, or the
benchmark's gold answer, and do not invent facts or citations. If the available
evidence does not support a decisive answer, choose Uncertain rather than
guessing.

Return exactly one JSON object and nothing else: no reasoning, commentary,
headers, or Markdown fences before or after it. Put any reasoning inside
"rationale" -- do not think out loud first. It must have this shape:

{
  "verdict": "Yes | No | Uncertain | Re-run | Human-review",
  "confidence": 0.0,
  "rationale": "concise explanation grounded in the supplied documents",
  "evidence": [
    {
      "kind": "primary_literature | cif | calc_result | db_record",
      "ref": "DOI, source path, database id, or calculation id",
      "resolvable": true,
      "fulltext_read": true
    }
  ]
}

The confidence must be a number from 0 through 1. Do not add keys outside this
schema. Evidence must describe only sources actually present in the supplied
context. Keep "rationale" to at most 3-4 sentences -- state the deciding fact
and its source, not a full restatement of the source text. A long rationale
risks being cut off before the JSON object closes, which makes the whole
response unusable.
\end{verbatim}
\endgroup

The item-level user message is:

\begingroup\scriptsize
\begin{verbatim}
BENCHMARK ITEM: {item_id}
TASK: {T-MOF-2|T-MOF-3}

QUESTION
{question}

Use every attached source document. Return the required JSON object now.

{source-paper PDF attachments and non-PDF source text blocks}
\end{verbatim}
\endgroup

\subsubsection{\textsc{MOF-Verify} procedures (T-MOF-1--4)}

The harness does not expose one shared language-model prompt across all task
families.  It routes each claim to the components in
Table~\ref{tab:failure-map}, and uses deterministic comparison whenever a
typed record is available.

\paragraph{Structural grounding (T-MOF-1).}
No language-model prompt is used.  The claim and item identifier are parsed;
\textbf{IR} resolves material identity, \textbf{SE} acquires and validates the
trusted deposited structure, and \textbf{VR} deterministically compares the
claim with the grounded record.  Gold labels and control metadata are not
provided to these components.

\paragraph{Synthesis-condition verification (T-MOF-2).}
\textbf{LE} checks the claimed synthesis conditions against the cited paper's
main text and supporting information and retains relevant source passages.

\paragraph{Evidence-sufficiency verification (T-MOF-3).}
\textbf{ES} assesses whether the paper's main text and supporting information
provide sufficient evidence to settle the claim; \textbf{VR} uses the resulting
evidence to assign the verdict. The paper-reading prompt used for T-MOF-2
and T-MOF-3 is shown below:

\begingroup\scriptsize
\begin{verbatim}
You are verifying a single factual claim about ONE chemistry paper. Judge ONLY
from the paper text below -- never from outside knowledge. Decide:
  Yes       -- the paper clearly SUPPORTS the claim.
  No        -- the paper clearly CONTRADICTS/refutes the claim.
  Uncertain -- the paper is silent, ambiguous, reports a different sample, or
               the claim needs a value/comparison the paper does not make.
A decisive Yes/No MUST be backed by an EXACT verbatim quote copied from the
paper. If you cannot quote support, answer Uncertain. When in doubt, answer
Uncertain -- a wrong Yes/No is worse than an abstain.
Respond as STRICT JSON only: {"verdict": "Yes|No|Uncertain",
"evidence_quote": "<verbatim from the paper, or empty>",
"rationale": "<one sentence>"}

=== PAPER ===
{paper_text, max 60,000 chars}

=== CLAIM ===
{question}
\end{verbatim}
\endgroup

\paragraph{MLIP-based computational verification (T-MOF-4).}
No language-model prompt is used.  \textbf{CE} parses the claimed calculation
and convergence controls, executes the SevenNet-based relaxation, and records
the calculator and input fingerprints.  \textbf{VR} then derives the verdict
deterministically from the typed computational result. This is the Track-4 arm; a separate Closed-book baseline is reported, while RAG and Oracle are inapplicable.

\subsection{Benchmark construction and implementation details}
\label{sec:implementation-details}

\subsubsection{No-tool answerability and benchmark strata}
We measured no-tool answerability with a fixed GPT-5.4 baseline using three
responses per item. Each response received 1 for an exact match to the reference
verdict and 0 otherwise; the item score was their mean. Items scoring at most
0.4 were assigned to the lower-answerability stratum, stored as
\texttt{calc\_required}, while higher-scoring items were retained as
\texttt{lookup}. The former is a stratum name, not a claim that every such item
requires an MLIP calculation. The lower-answerability/lookup counts are
244/32 for T-MOF-1, 768/101 for T-MOF-2, and 36/81 for T-MOF-3, giving the
276, 869, and 117 items in Table~\ref{tab:dataset}. The reported task-level
results pool both strata. This construction does not require every evaluated
model to fail on every item; even one correct response out of three lies below
the 0.4 threshold.

\subsubsection{Recorded paper-reading requests}
For T-MOF-2 and T-MOF-3 in the harness condition, paper reading used the backbone
named in each result row. Recorded OpenRouter client requests specified
\texttt{temperature=0} and \texttt{top\_p=1} for nine backbones; Gemini batch
requests omitted these fields. The requested reasoning effort was
\texttt{none} for those nine backbones and \texttt{low} for Gemini. All these
requests specified \texttt{max\_tokens=1024} and \texttt{seed=20260819}.
These are client-requested parameters, not independently verified
provider-effective decoding settings. The preserved item traces contain one
or two attempts, rather than establishing a single successful call for every
item. These settings describe the harness paper-reading calls, not every
evaluation arm. The paper-reading prompt and its 60,000-character input limit
are given in Appendix~\ref{sec:evaluation-prompts}.

\subsubsection{Plain-RAG retrieval and input assembly}
Each question is paired with up to eight ranked source-text excerpts.
Each excerpt carries its chunk identifier, source identifier, main-text/SI
designation, and location. The excerpts are inserted into the retrieval-enabled
user-message template in Appendix~\ref{sec:evaluation-prompts}; this condition
does not supply the structured harness records or an oracle evidence packet.
The two archived retrieval packs share a frozen corpus of 1,252 extracted
documents (700 main-text documents and 552 supporting-information documents),
split into 34,957 chunks. We used \texttt{Qwen/Qwen3-Embedding-4B}, revision
\nolinkurl{5cf2132abc99cad020ac570b19d031efec650f2b}, for corpus and query
embeddings. Documents were split into 512-token windows with 64-token overlap
using the embedding model's tokenizer. Normalized 2,560-dimensional last-token embeddings were ranked
by exact cosine similarity using NumPy, retaining the top eight chunks without
reranking or query expansion. Each selected chunk was supplied in full, without
additional post-retrieval truncation. The corpus snapshot identifier is
\nolinkurl{cadaa9d012dfd01628aba0e3145e6cef52c59ae783c9e2d4373d1efaf73e2044}.
For 144 of the 276 T-MOF-1 items, the archived RAG queries use historical
wording that differs from the later canonical item text; these archived runs
are retained in the reported results.

\subsubsection{Structured synthesis resources and reference annotations}
The T-MOF-2 harness uses a hash-pinned snapshot of structured synthesis records
containing reported conditions, supporting quotations, and document provenance.
The snapshot contains 1,786 records from 383 DOIs and is stored separately from
the benchmark reference-verdict annotations. It is a partial extraction resource,
not a claim of complete or independently domain-reviewed coverage. Plain-RAG
receives retrieved source-text excerpts, whereas the harness additionally has
access to these structured records. Separate storage does not establish that
the records were curated independently of the benchmark reference annotations.

\end{document}